\documentclass[letterpaper, 10 pt, conference]{ieeeconf}
\usepackage{graphicx}
\usepackage{booktabs} 
\usepackage{ragged2e} 
\usepackage{dsfont}   
\usepackage{makecell}
\usepackage{mathtools}
\usepackage{hyperref}  
\usepackage{nth}
\hypersetup{
    colorlinks=true,
    linkcolor=blue,
    filecolor=magenta,      
    urlcolor=black,
    pdfpagemode=FullScreen,
    }

\usepackage{amsmath}   
\graphicspath{{visuals/}}
\usepackage[backend=biber,style=numeric,natbib=true,sorting=none,giveninits=true,maxbibnames=9]{biblatex}
\title{\LARGE \bf
Hybrid Control for Robotic Nut Tightening Task
}
\author{Dmitri Kovalenko\\\small{IEEE Member, Helsinki, Finland}}
\begin{document}
\maketitle
\thispagestyle{empty}
\pagestyle{empty}
\begin{abstract}
An autonomous robotic nut tightening system for a serial manipulator equipped with a parallel gripper is proposed.
The system features a hierarchical motion-primitive-based planner and a control-switching scheme that alternates between force and position control.
Extensive simulations demonstrate the system's robustness to variance in initial conditions.
Additionally, the proposed controller tightens threaded screws $14 \%$ faster than the baseline while applying $40$ times less contact force on manipulands.
For the benefit of the research community, the system's implementation is open-sourced.
\end{abstract}

\relax\par\normalfont\normalsize

\section{Introduction}
The robotics research community is working towards autonomous robotic systems that
    could multiply the productivity of every individual and organisation.
Presently versatility of the planet's best manipulator
  (a human's hand controlled by human's mind \& reflex~\cite{elliott1984classification}) is unparalleled by any autonomous robotic system.
There is an argument \cite{ding2021design} that traditional research problems in manipulation, such as
    pick-and-place, peg-in-hole, are not sufficient for advancing the field toward more challenging practical applications,
    which require mastering sophisticated contact interactions between the autonomous system and its manipulands.

The present work aims to address these challenging practical applications by exploring a novel hybrid control scheme for a standard robot used in manipulation research -
  a serial manipulator with a parallel gripper.
The specific manipulation task considered is nut tightening.
This task is significant for two reasons.
First, the assembly of threaded fasteners has widespread applications, ranging from heavy industries to home improvement \cite{gautam2020collaborative}.
Second, nut tightening is inherently complex~\cite{noda2019}, as it cannot be easily reduced to classic motion primitives like pick-and-place.

A large body of literature covers fastening manipulation tasks, with 
    early contributions \cite{yoshimi1997integrating} emerging during the dawn of the ubiquitous computing era.
These tasks are often formulated in terms of impedance control~\cite{hogan1984impedance} or hybrid control~\cite{raibert1981hybrid}.
The valve-turning task is closely related to fastening, and therefore, works addressing the former are also considered here.
Manipulation research is equally active for UAV platforms; however,
    due to significant differences in implementation, most of this research falls outside the scope of the present work.
An exception is~\cite{ding2021design}, where a robust controller enables screwdriving in UAV's lateral plane.
Dynamic movement primitives are used in~\cite{carrera2015learning} to plan autonomous valve turning by an underwater robot.
Ajoudani et al.~\cite{ajoudani2014manipulation} employ impedance control to manipulate a valve in a dual-arm setup.
Shauri et al.~\cite{shauri2012sensor} proposed a dual-arm manipulation framework that not only performs the fastening task,
    but also addresses object pickup and the small-clearance peg-hole subproblem (fitting a nut onto a bolt).
The screwing task is explored in a collaborative setting in~\cite{villa2022contact}, where an operator can guide the manipulator during the process.
The proposed controller in~\cite{villa2022contact} demonstrated superior accommodation of the operator's interventions compared to classical impedance control.
The authors in~\cite{an2024robot} introduce a novel perception modality for controlling a screwing task using a tactile sensor,
while~\cite{golani2024robotic} presents the estimation framework that derives gripper-valve misalignment from sensed torques.

Unlike the force-controlled screwdriving framework in~\cite{tang2023robotic}, the present work does not consider tool use.
Zhang el al.~\cite{zhang2024autoregressive} propose an autoregressive deep-learning model for manipulation planning, which consistently achieves successful nut tightening in real-world experiments.
Meanwhile, the authors in~\cite{fujioka2022hex} aim to bridge the productivity gap in collaborative robotics by incorporating high-frequency camera feedback into manipulation control.

\begin{figure}[t!]
\includegraphics[width=.4\textwidth]{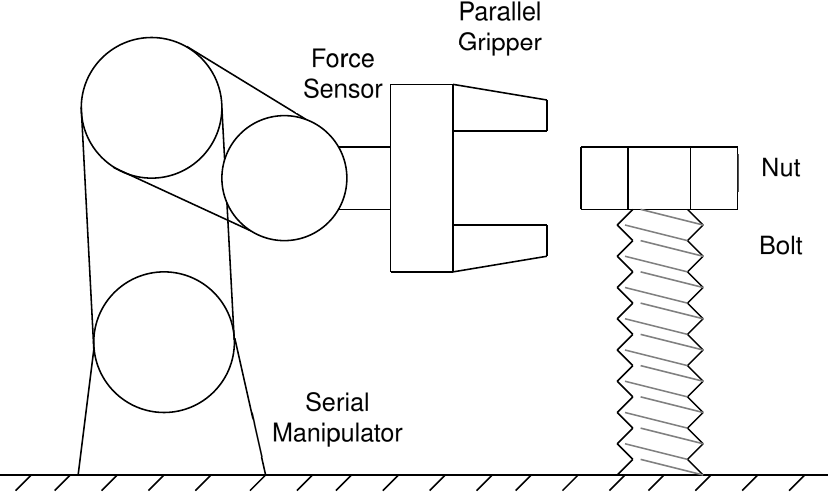}
\caption{
  A schematic depiction of the considered task.
  An autonomous manipulation system that uses a parallel gripper and signals from F/T sensor to tighten a nut onto a bolt is proposed.
}
\label{vis:scheme}
\vspace{-1.0em}
\end{figure}
Manipulation research which considers other tasks besides fastening is still relevant for the present inquiry;
    a survey by Suomalainen et al.~\cite{suomalainen2022survey} presents autonomous systems with such contact-rich capabilities as wiping, scooping and part assembly.
An elastic end-effector is used jointly with torque-position control to demonstrate~\cite{suh2022seed} successful tool use for wiping.
Hou et al.~\cite{hou2019robust} present a hybrid force-velocity control framework that models both manipulands and the robot throughout contact-rich phases of their trajectories.

The key contributions of this work are:
  a) an autonomous robotic manipulation system for nut tightening with a serial manipulator;
  b) a hierarchical motion-primitive-based planning framework;
  c) a robust hybrid control scheme with a secondary objective.

\begin{figure}
\includegraphics[width=.48\textwidth]{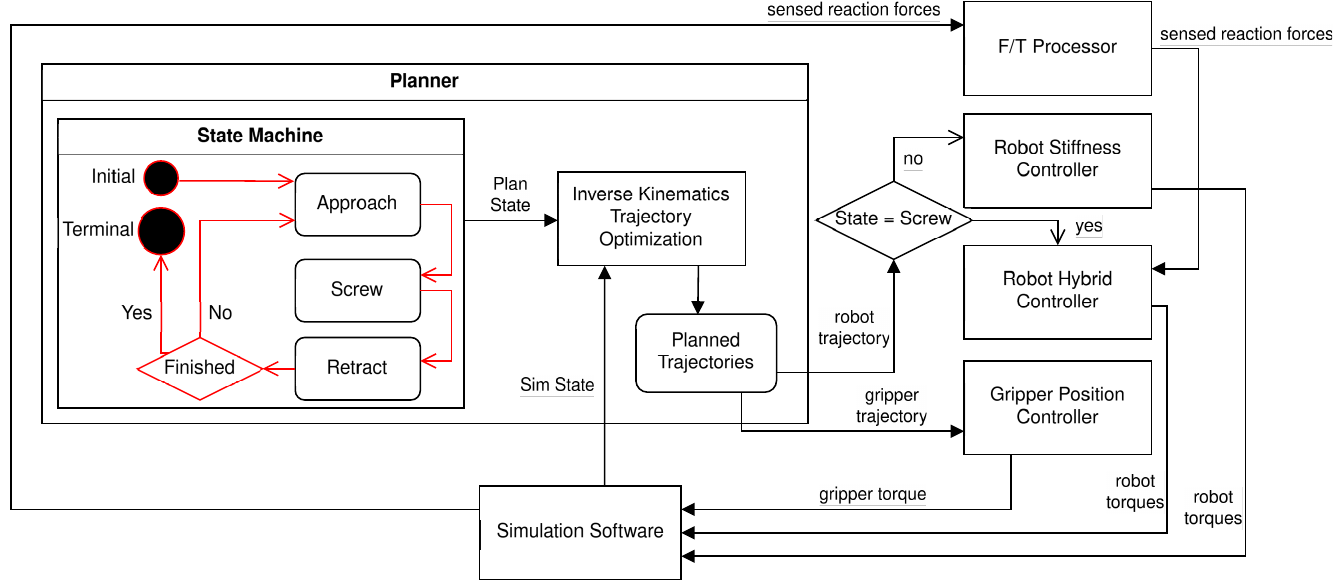}
\caption{
  The proposed autonomous manipulation system architecture for nut tightening task.
  The planner iterates through the \textit{approach}, \textit{screw}, \textit{retract} stages to guide the robot's interaction with the manipuland.
  Hybrid control is applied during the \textit{screw} stage and stiffness control is applied otherwise.
}
\label{vis:architecture}
\end{figure}

The author's goal is to develop such an autonomous system and experimentally verify it, in order to unlock author's future research in contact-rich manipulation.
The proposed system was evaluated via simulation, and the visual perception part of the robot's autonomy stack was de-emphasized.
This is the reason for absence of the visual perception component in the system's architecture diagram (Fig.~\ref{vis:architecture}).
The remaining components in the robot's autonomy stack (the planner and control) rely on a perfect oracle provided by the simulation software for geometric perception data.
A detailed presentation of said components follows.

\section{Planning Framework}

The planner employed in the proposed system is hierarchical in that,
    firstly, a sequence of task space "keyframes" is constructed which the robot's gripper needs to traverse (\cite{lavalle2006planning},~Ch.7), then
    secondly, an optimisation problem is solved in order to obtain a continuous trajectory for the controller to execute.
The latter part of the planning computation is uniform across the manipulation stages, while the former part is stage-specific and exploits properties of the task.

\subsection{Keyframe Planning}\label{sec:kf-plan}

During the keyframe planning the \textit{approach} and \textit{retract} stages mirror one another temporally,
    and consist of three keyframes (an example is shown in Fig.~\ref{vis:plan-knots}.a,c).
The approach stage plan includes the initial keyframe, the pre-grasp keyframe and the grasp keyframe. The retract plan is the same but in reverse.
The initial keyframe is determined trivially.
The selection of grasp keyframe relies on the dedicated procedure visualised in Fig.~\ref{vis:grip-selection}.
The need for intermediate keyframes "pre-grasp" and "post-grasp" is due to avoiding spurious contact with the manipuland.
The robot's gripper must approach and retract radially relative to the bolt's center to achieve a stable grasp.
A hexagonal manipuland offers six distinct directions for grasping.
The planning framework presently retrieves these directions from the visual perception oracle available in the simulation software.
Then such a direction is selected which is the one closest to
    the present manipulator's inner space configuration ($q_{\text{now}} \in \mathds{R}^7$)
    provided it is within the manipulator's joint limits.
\setlength{\tabcolsep}{0.25em}
{
\begin{figure}[t!]
  \begin{tabular}{cc}
    \includegraphics[width=.19\textwidth]{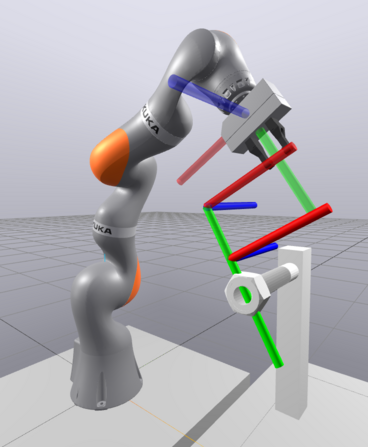} &
    \includegraphics[width=.19\textwidth]{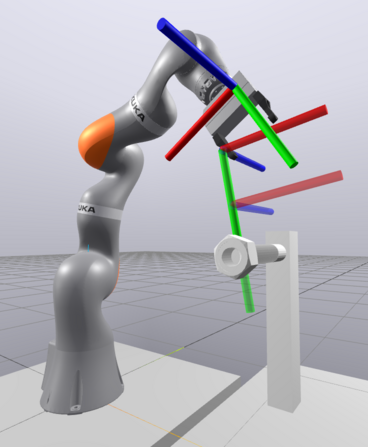} \\
    \small{a)} & \small{c)} \\
    \multicolumn{2}{c}{\includegraphics[width=.25\textwidth]{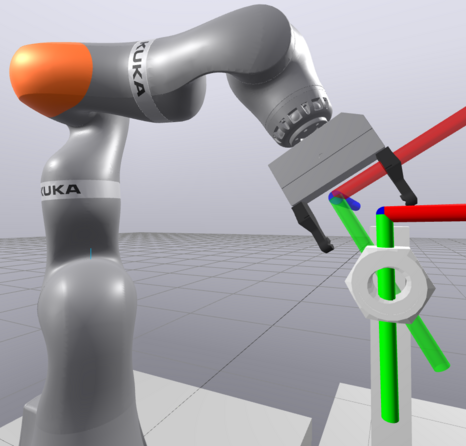}} \\
    \multicolumn{2}{c}{\small{ b) }} \\
  \end{tabular}
\caption{
  The planning framework models explicitly the three stages of nut tightening task: a) approach, b) screw, c) retract.
  Note: the forward direction of the gripper (co-axial with the major axes of the gripper's fingers) coincides with the Y-axis of the gripper's model (green).
  The opacity of the keyframes encodes the precedence information: lower-opacity keyframes must be reached before higher-opacity ones.
  30$^{\circ}$ turn plan is shown in b).
}
\label{vis:plan-knots}
\end{figure}
}
Finally, the maximal rotation around the manipuland that does not result in robot's kinematic singularity
  informs the keyframe selection for the \textit{screw} stage plan (Fig.~\ref{vis:plan-knots}.b).

\subsection{Trajectory Planning}
At the time of the stage transition (see the planner's state machine in Fig.~\ref{vis:architecture}),
    the trajectory optimisation is conducted.
It solves for manipulator's inner space configuration $q$ at the time of the new stage, provided $q_0$, which is the previous stage configuration or a nominal one.
\begin{equation} \label{eq:trajopt}
  \begin{split}
           \hat{q} & \leftarrow q_0 \\
    \min_{\hat{q}} & ( \hat{q} \cdot  q_0 )^2 \text{ s.t.} \\
           \| {T_{\text{keyframe}}}^W_G & - T^W_G(\hat{q}) \| < [0.01, 0.01, 0.01]^T \\
             -0.5^{\circ} < & {[{R_{\text{keyframe}}}^W_G]}^{-1} \cdot R^W_G(\hat{q}) < 0.5^{\circ}
  \end{split}
\end{equation}
$T \in R^3$, $R \in \text{SO}3$ in (\ref{eq:trajopt}) stand for the translation and attitude of bodies, e.g.: gripper (G), expressed in the world frame (W).
All keyframe-subscripted quantities are input to this optimisation and are estimated before (Sec.~\ref{sec:kf-plan}).
The closest to $q_0$ configuration of $\hat{q}$ is optimised for,
  while guaranteeing the gripper's pose being within the tolerance limits of the keyframe pose.
The optimisation runs SNOPT~\cite{gill2018user} that is bundled with the simulation software.
A sequence of these optimisations over the plan's keyframes generates the sequence of manipulator's generalised coordinate vectors.

A trajectory optimisation failure constitutes the termination condition for the autonomous system.
Conversely, upon the successful termination of the optimisation,
  a continuous inner space trajectory is constructed for the manipulator using the first-order hold and
  is communicated to the controller via the message passing system.

The above discussion did not cover the planning for gripper trajectories because of its triviality in the proposed planning framework.
The keyframe plans admit dedicated time windows before and after nut tightening interactions for gripper's fingers to open and close.
The manipulator is kept stationary during fingers actuation.
Provided the actuation time windows, a 1D gripper position trajectory is generated and communicated to a dedicated controller.

The planner executes this policy greedily without feedback, except reacting to the termination conditions, which could be \textit{normal} or \textit{exceptional}.
Either the total amount of commanded nut tightening manipulations is completed successfully, or
    the trajectory optimisation has diverged (\textit{exceptional}).
The experimental evaluation of the proposed system (Sec.~\ref{sec:experiments}) has examples of both,
    while the following section details the control framework.
\setlength{\tabcolsep}{0.25em}
{
\begin{figure}
  \begin{tabular}{cc}
    \includegraphics[width=.19\textwidth]{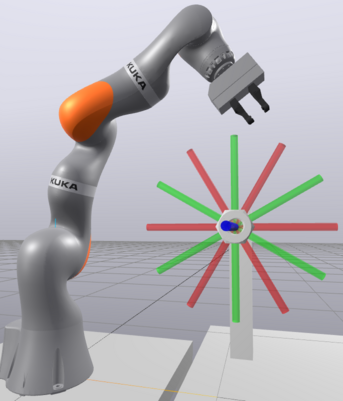} &
    \includegraphics[width=.19\textwidth]{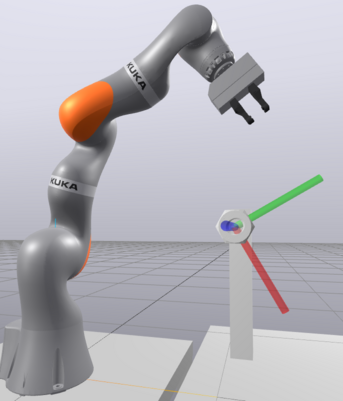} \\
    \small{ a) } & \small{ b) } \\
  \end{tabular}
\caption{
    The grasp selection problem is solved during the keyframe planning for the \textit{screw} stage.
    a) A hexagonal manipuland allows for six distinct stable grasps. b) The grasp that is the most advantageous kinematically is selected.
}
\label{vis:grip-selection}
\end{figure}
}

\section{Control Framework}
The goal of the controller components in Fig.~\ref{vis:architecture} is to execute the trajectories obtained from the planner.
The central focus of this work is to study the impact that varying control schemes have on the autonomous robotic system.
Hereafter, the control schemes which were experimentally evaluated are explained in detail.

\subsection{Stiffness Controller}\label{sec:stiff}
This position control scheme is used throughout \textit{approach}, \textit{retract} stages of manipulation and
    is dedicated to tracking the manipulator's inner space trajectories $Q_d^{t \in \{T_n \ldots T_m \}}$,
    with $T_n$, $T_m$ representing the start and end times of the trajectories.
Here and in the following, the variables subscripted with \textit{d} denote \text{desired} values, while the subscript \textit{s}, respectively, means \textit{sensed} (i.e. measured).
The manipulator's inner space dynamics in the absence of contact forces are formulated below.
This implementation is informed by~\cite{lewis2003robot}.

\begin{equation} \label{eq:dynamics}
\tau = M(q) \ddot{q} + C(q, \dot{q}) \dot{q} + g(q)
\end{equation}

$M$ is the mass matrix, $C$ is the Coriolis and gyroscopic effects term, $g$ is the gravity term.
Then, the commanded robot torques $\tau \in \mathds{R}^7$ are:
$\tau = \tau_{0} + PD(q, \dot{q})$, where $PD(q, \dot{q}) = M(q)(K_p \cdot (q_d -q) + K_d \cdot (\dot{q_d} - \dot{q}))$,
and $\tau_{0}$ is a stabilizing term for withstanding the gravity force, obtained from (\ref{eq:dynamics}).

\subsection{Hybrid Controller}
Aside from the inherent distinction of controlling the force instead of velocity along a certain dimension,
    the proposed hybrid controller differs from the one described above in that it is formulated in task space~\cite{lewis2003robot}.
This is because the force-controlled dimension depends on the pose of the manipuland.
The manipulator's inner trajectory $Q_d$ is also converted
    into a gripper pose trajectory $X_d$ (where each datum is a SE3 pose, $x^T = (\alpha, \beta, \gamma, t_x, t_y, t_z)$)
    for controlling the dimensions orthogonal to the one being force-controlled.
An additional input to this controller is the sensed force wrench $\rho^T = (M_x, M_y, M_z, f_x, f_y, f_z)$,
    which is projected into the manipuland's coordinate frame by the F/T processor.
Following~\cite{zhang1985hybrid}, the control is:
\begin{subequations} \label{eq:hybrid}
\begin{align}
  \tau = & \tau_0 \label{eq:hybrid-a} \\
       + & K_p \cdot J^T(x_d - x)_{4,6} + K_d \cdot J^T(\dot{x_d} - \dot{x})_{4,6} \label{eq:hybrid-b} \\
       + & K^f_p \cdot J^T(\rho_d - \rho_s)_{2} + K^f_d \cdot J^T(\dot{x_d} - \dot{x})_{2} \label{eq:hybrid-c} \\
       + & \epsilon (I - J^{\dagger} J) [ K^{2\text{nd}}_p \cdot (q_{0} - q) + K^{2\text{nd}}_d \cdot \dot{q} \label{eq:hybrid-d} ]
\end{align}
\end{subequations}
In (\ref{eq:hybrid}) $J$ is the $6 \times 7$ manipulator's spatial velocity Jacobian with respect to the manipuland's coordinate frame. $J^{\dagger}$ is its pseudoinverse.
Note, that error term vectors are zero-padded to match the dimension of $J^T$.
The first two rows (\ref{eq:hybrid-a}, \ref{eq:hybrid-b}) parallel the terms present in stiffness control (Sec.~\ref{sec:stiff}),
    with the distinction that position control is applied only along maipuland's $t_x$, $t_z$ axes.
The third row (\ref{eq:hybrid-c}) describes the moment of force control around manipuland's major axis ($r_y$, $\beta$).
${\rho_d}_2 = M^{\text{desired}}_y = 0.2\ \text{N} \cdot \text{m}$.
The pitch angle rate is used in the differential term because the measured force signal is not differentiable.
The last row (\ref{eq:hybrid-d}) implements null space projection for the secondary task ($\epsilon = 10^{-2}$),
  which is the joint centering objective towards the nominal configuration $q_0$.

\section{Experiments}\label{sec:experiments}

\setlength{\tabcolsep}{0.15em}
{
\begin{figure}[t]
  \centering
  \footnotesize{
  \begin{tabular}{ c c c c }
    \includegraphics[width=.1\textwidth]{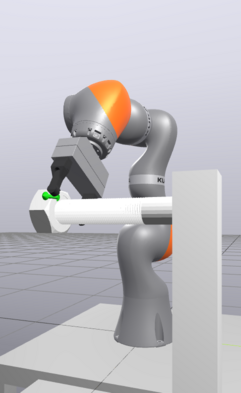} &
    \includegraphics[width=.1\textwidth]{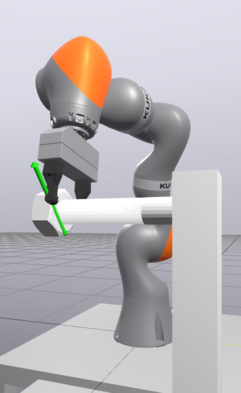} &
    \includegraphics[width=.1\textwidth]{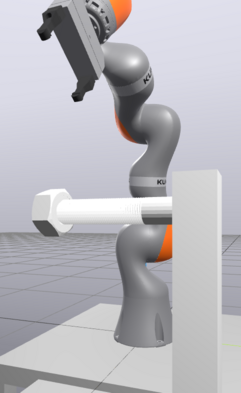} &
    \includegraphics[width=.1\textwidth]{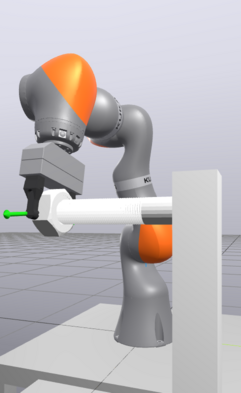} \\
    \multicolumn{3}{ c }{\nth{1} turn: approach, screw, retract} & \nth{2} turn: approach\\
    \includegraphics[width=.1\textwidth]{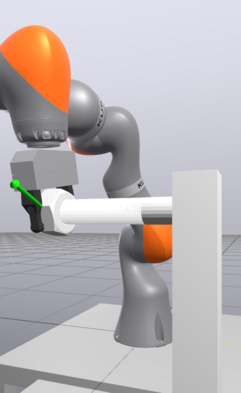} &
    \includegraphics[width=.1\textwidth]{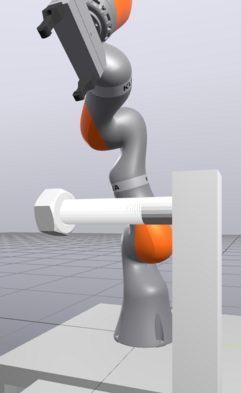} &
    \includegraphics[width=.1\textwidth]{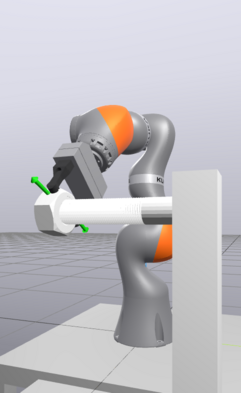} &
    \includegraphics[width=.1\textwidth]{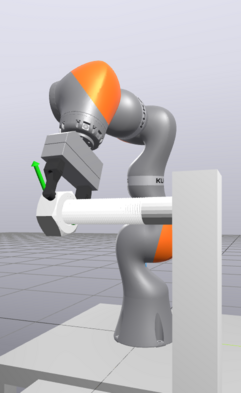} \\
    \multicolumn{2}{ c }{\nth{2} turn: screw, retract} & \multicolumn{2}{ c }{\nth{3} turn: approach, screw} \\
  \end{tabular}
  }
\caption{
  A visualisation of simulated manipulation sequence by the proposed system.
  The robot completes three nut tightening sequences, with grasp adjustments between the screwing interactions.
  Each sequence includes the \textit{approach}, \textit{screw}, \textit{retract} stages.
  Green arrows indicate the directions and relative magnitudes of the contact forces.
  The initial and terminal states of the system are omitted for brevity.
}
\label{vis:key-stages}
\end{figure}
}

The proposed nut tightening manipulation system was evaluated in simulation using the MIT Drake software~\cite{drake}.
The manipulator used in the testing scenario is a collaborative robot 7 DoF KUKA iiwa with the parallel gripper by Schunk.
The scenario involves the robot starting in a nominal position, then reaching towards the manipuland, turning it, and then returning back to the nominal position.
The manipulating sequence is repeated several times, as a nut progressively gets tightened onto a bolt.
The main stages along this sequence are imaged in Fig.~\ref{vis:key-stages}, where the robot completes three turning interactions.
The simulation is run in discrete mode with an integration step of $10^{-4}$~sec. The bolt's thread pitch is $8$~mm.
The control parameters include: a) $K_p = 100$, $K_d = 20$ (stiffness control),
                                b) $K^f_p = 0.2$, $K^f_d = 10$ (force control),
                                c) $K^{2\text{nd}}_p = 100$, $K^{2\text{nd}}_d = 20$ (the secondary objective, position control).
When stated separately in an experiment description, another set of parameters is randomised in the simulation for statistical analysis.
These include the initial world poses of the gripper and manipuland.
Instantaneous states of the simulated world are accumulated across all experiments for the quantitative evaluation that follows.

\subsection{Ablation Study}\label{sec:abla}
The proposed manipulation system features a novel hybrid controller.
Therefore, a comparison with a variant of the system where only the baseline controller is used is required.
The proposed system uses the stiffness control (Sec.~\ref{sec:stiff}) for non-contact portions of control sequence.
This same controller is employed throughout the simulation to obtain a baseline result,
    whereas the full result is obtained when the control is switching to hybrid as illustrated in Fig.~\ref{vis:architecture}.
Two versions of the manipulation system (baseline and full) share the planning framework in its entirety, thus planned manipulation sequences will have similar properties.
Notably, these sequences will be of the same length.

\begin{figure}
\includegraphics[width=.48\textwidth]{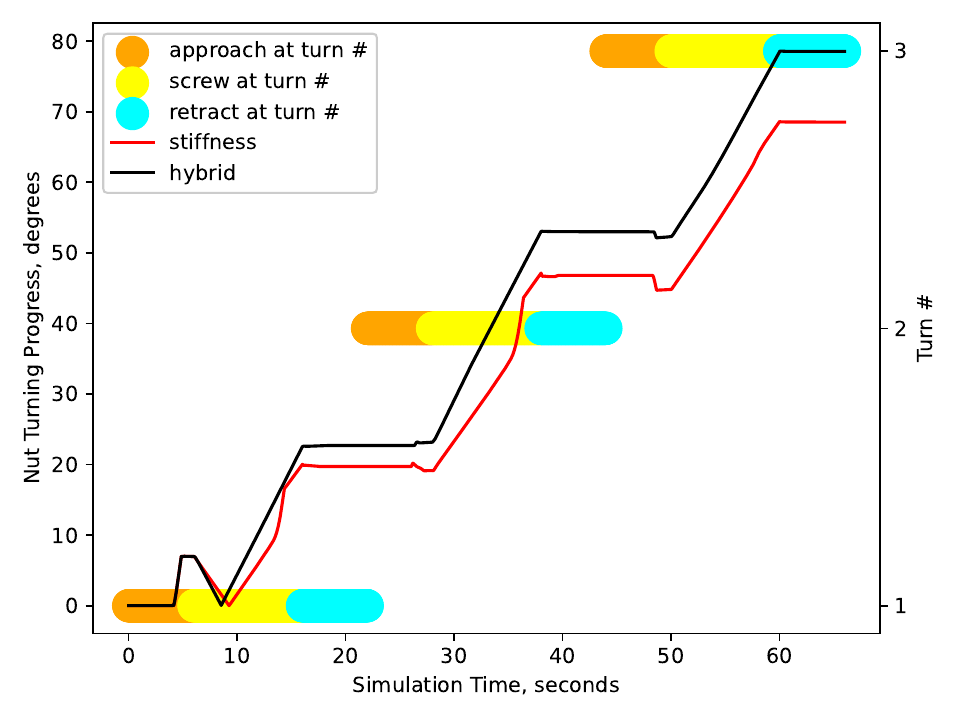}
\caption{
    \textit{Higher is better}.
    The changes in the nut's pitch angle for both the baseline (stiffness) and the proposed (hybrid)
        manipulation systems are compared during simulations, starting from the same initial configuration.
    The proposed system demonstrates a $14.5 \%$ greater progress in reference to the baseline.
    Notably, the baseline plot shows undesirable jaggedness, suggesting excessive contact force.
    The color bars highlight the stages of the manipulation sequence as shown in Fig.\ref{vis:key-stages}.
}
\label{vis:turning-progress}
\end{figure}

The efficiency is then judged by the amount of movement generated in the manipuland by the robot during tests,
    all of which take the same amount of simulation time.
This comparison is presented in Fig.~\ref{vis:turning-progress},
    where the color bars mirror the \textit{approach}-\textit{turn}-\textit{retract} stages of Fig.~\ref{vis:key-stages} while overlaying the simulation timespan.
The proposed manipulation system demonstrates a $14.5 \%$ efficiency advantage over the baseline.
The smoothness or jaggedness of manipuland's pitch angle trajectory indicates whether excessive force was ever commanded by the control.
The pitch angle has a smoother curve in the proposed system.

\begin{figure}
\includegraphics[width=.48\textwidth]{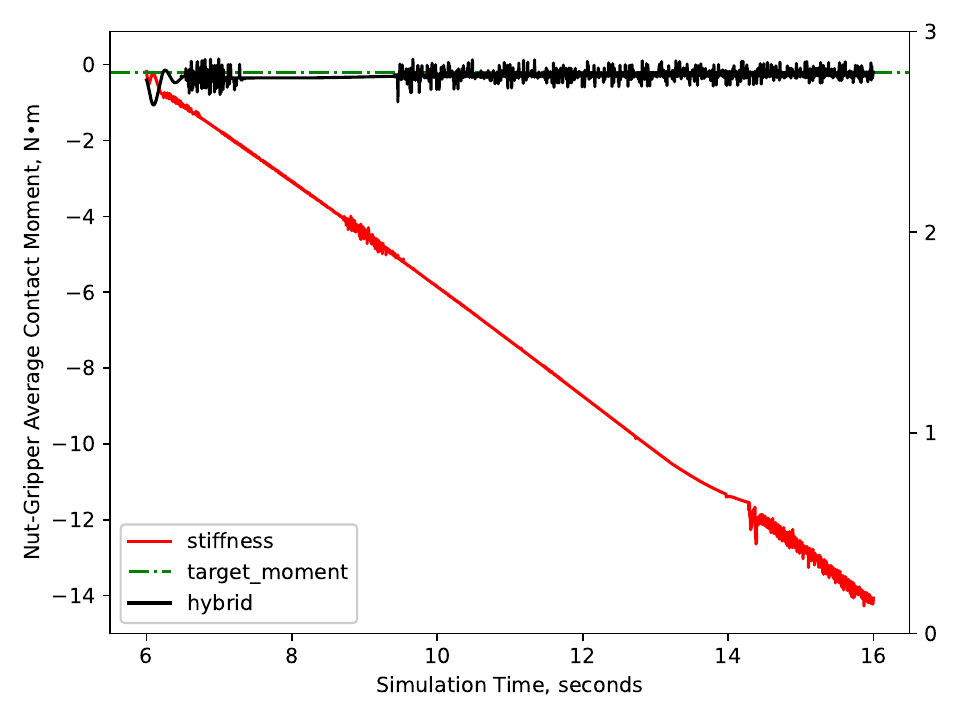}
\caption{
    \textit{Lower magnitude is better}.
    The averaged moments of contact force $M_y$ between the gripper and the manipuland for the baseline (stiffness) and the proposed (hybrid) manipulation systems.
    The green horizontal line represents the commanded moment for the hybrid controller.
    The baseline controller exerts excessive force on the manipuland during the screwing stage.
}
\label{vis:moments-comp}
\end{figure}

In order to highlight this distinction further, the gripper-manipuland contact wrench was recorded from the simulation's states for both systems under study.
Fig.~\ref{vis:moments-comp} plots the force profiles, averaged over three consecutive turning motions within the same manipulation sequence.
Among all the dimensions of the contact wrench, the moment of force, which is co-axial with the bolt's major axis~($M_y$), is of primary interest.
This exact dimension is used in Fig.~\ref{vis:moments-comp},
    and allows to observe that the proposed system shows no trend in commanded force, whereas the baseline applies a linearly growing force to the manipuland,
    while maintaining near-constant velocity (as demonstrated in Fig.~\ref{vis:turning-progress}).

\begin{figure}
\centering
\begin{tabular}{c}
  \includegraphics[width=.48\textwidth]{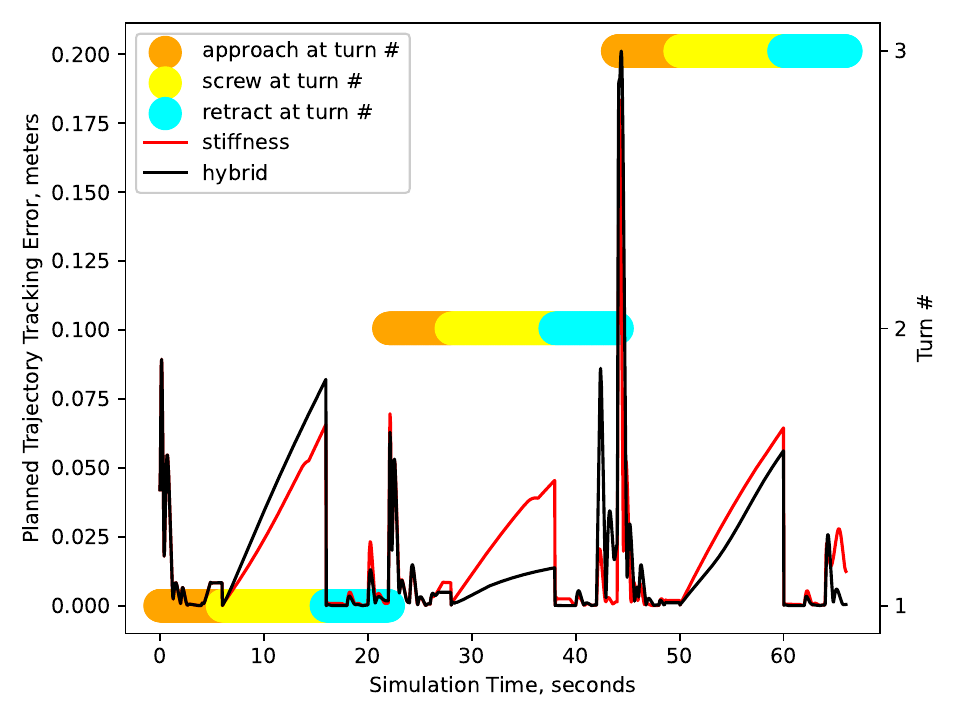} \\
  \small{a)}\\
  \includegraphics[width=.48\textwidth]{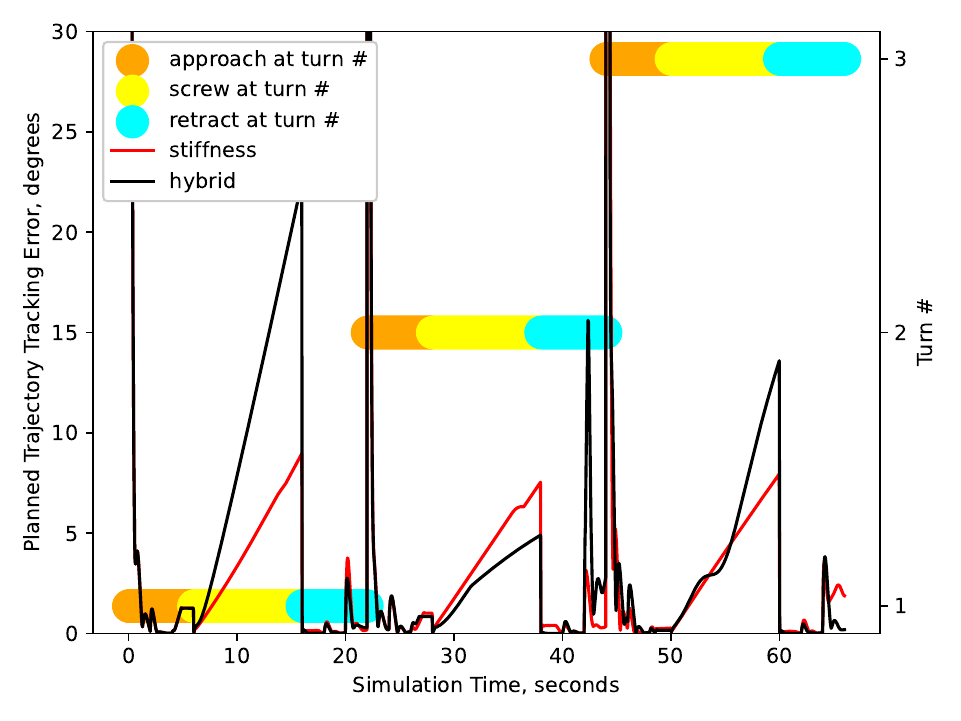} \\
  \small{b)}\\
\end{tabular}
\caption{
  \textit{Lower is better}.
  The planned trajectory tracking errors in translation (a) and rotation (b) for the baseline (stiffness) and the proposed (hybrid) manipulation systems.
  The proposed system has higher error during the screw stages because the controller executes a faster trajectory than the planner suggested.
  For a discussion on error spikes during contact-free stages, refer to Sec.~\ref{sec:abla}.
}
\label{vis:tracking-error}
\vspace{-1.0em}
\end{figure}

The dimensions which are not force-controlled are position-controlled in the proposed system.
To further verify it, the system is compared to the baseline in terms of the tracking error (Fig.~\ref{vis:tracking-error}).
The proposed system appears to have progressively higher error during the turning stages; however, this is a consequence of the hybrid control.
The commanded moment of force allows the manipulator to tighten the nut faster than it was specified in the planned trajectory.
The reason for the increasing errors in the baseline system during the turning stages is different:
this is due to the planned trajectory being wrapped around the manipuland without explicitly accounting for the interaction forces.
As expected, the \textit{approach} and \textit{retract} stages do not differ significantly between two systems.
The contact-free parts of the manipulation sequences are controlled by the same position control.
The spikes in error plots during these stages, as seen in Fig.~\ref{vis:tracking-error}, are caused by
    discontinuities at the boundaries of the trajectories produced for adjacent stages by the planning framework.
Unrelated to this ablation study, the issue is considered further in Sec.\ref{sec:conclusion}.

\subsection{Evaluation}
The holistic evaluation of the proposed system concludes the experiments section.
This test examines its robustness to variance in the initial configuration: translations of both the gripper and the manipuland are randomised.
The random distributions have progressively larger supports to capture extreme cases in the system's behavior.
Over 30 random trials were conducted, with the statistics summarized in Tab.~\ref{tab:robustness}.
A trial is considered successful if at least one turning interaction is fully completed during the simulation.
A randomised position disturbance ($d \in \mathds{R}^3$) is added to the initial pose of either the manipuland or the gripper.
In the case of the gripper, inverse kinematics is used to compute a configuration that satisfies the new initial pose.
The orthogonal axes of $d$ are randomised independently via a uniform distribution: $d \sim U[-l \ldots l]$, where $l$ stands for the support's boundaries.
The tests with varying parameter $l$ are listed in Tab.~\ref{tab:robustness}
    and demonstrate that the proposed system is robust even under significant disturbances.
The larger displacement simulations that did not succeed failed due to a limitation in the planning framework.
In no case did the system fail to establish a stable grasp or execute a fastening turn,
    but the planner's greedy policy caused the manipulator to translate into configurations
    where the \textit{retract} stage plan fails to converge.
This limitation is discussed in the following section.

\begin{table}
\caption{
  The manipulation robustness in a nut tightening task
}
\centering
{
  \begin{tabular}{l|c|c}
  \toprule
    \makecell{Initialisation \\disturbance limit, m} & Trials amount & Success rate, $\%$ \\
  \midrule
    0.1   &  10  & 100 \\
    0.125 &  10  & 70 \\
    0.15  &  10  & 50 \\
  \bottomrule
  \end{tabular}\\
}
\vspace{0.4em}
\justifying{\footnotesize{
    The initial positions of the gripper and manipuland are disturbed by a random 3D offset,
        drawn from a uniform distribution within the specified limits.
    Notably, these offsets are relatively large, being $1.5$ times the length of the manipuland's diameter.
    The proposed manipulation system successfully completes the task in the majority of simulations.
    The typical failure mode is a denial from the planning framework at the \textit{retract} stage.
}}

\label{tab:robustness}
\vspace{-1.0em}
\end{table}

\section{Conclusions}\label{sec:conclusion}
The proposed robotic nut tightening system (a) is comprised of:
  b) motion-primitive-based planning framework operating in the task space, and
  c) hybrid controller that leverages sensed interaction forces to execute the contact-rich portion of the planned trajectory more efficiently.

The experimental evaluation revealed that this system is $14.5 \%$ faster in completing the objective compared to the baseline system,
  while being safer and more efficient by applying two orders of magnitude less contact force to the manipuland than the baseline system.

Both the planning and control components of the proposed system have low computational costs,
  consuming a negligible fraction of CPU resources compared to the simulation software with which they were run.

The system demonstrated high robustness to variance in initial configuration with a clear direction for further improvements.
One exisiting robustness bottleneck lies in the \textit{retract} motion primitive of the planning framework.
In future work, the tighter coupling between planning and control will mitigate the position tracking error spiking on switches to the \textit{retract} stage.
Further, the proposed system will obtain perception capabilities and enable experiments in a physically embodied setup.

The open source implementation of the system is made available at \textbf{\href{https://github.com/wf34/study_controllers}{https://github.com/wf34/study\_controllers}}.

\printbibliography

\end{document}